\pdfoutput=1

\documentclass[11pt]{article}

\usepackage[]{acl}

\usepackage{times}
\usepackage{latexsym}
\usepackage{pgfplots}
\usepackage{lipsum} 
\usepackage{multirow}
\usepackage{colortbl}
\usepackage{booktabs}
\usepackage{amssymb}

\usepackage{tikz}

\usepackage[draft]{todonotes}

\usepackage{array,booktabs,arydshln,xcolor}
\usepackage{verbatim}
\usepackage{url}
\usepackage{graphicx}

\definecolor{myred}{RGB}{180, 31, 39}
\definecolor{light-gray}{gray}{0.4}

\definecolor{lightyellow}{HTML}{eb9634}
\definecolor{lightred}{HTML}{FFCCCC}
\definecolor{lightpurple}{HTML}{CCCCFF}
\definecolor{lightblue}{rgb}{.90,.92,1}
\definecolor{lichen}{rgb}{.91,.95,0.83}

\usepackage{graphicx}
\usepackage{float}
\definecolor{hred}{RGB}{247,202,197}
\definecolor{horange}{RGB}{242,217,198}
\definecolor{hyellow}{RGB}{247,225,147}
\definecolor{hlgreen}{RGB}{200,207,141}
\definecolor{hdgreen}{RGB}{172,215,156}
\definecolor{hlblue}{RGB}{181,228,227}
\definecolor{hdblue}{RGB}{188,220,246}
\definecolor{hpurple}{RGB}{219,218,246}

\definecolor{hdblue}{RGB}{188,220,236}
\definecolor{hdblue2}{RGB}{188,220,236}

\definecolor{mygray}{HTML}{C0C0C0}

\usepackage{stfloats}
\usepackage[T1]{fontenc}

\usepackage[utf8]{inputenc}

\usepackage{microtype}

\title{\textsc{SciNLI}: A Corpus for Natural Language Inference on Scientific Text}

\author{Mobashir Sadat and Cornelia Caragea \\
Computer Science \\
  University of Illinois Chicago \\
  \texttt{msadat3@uic.edu} \mbox{    }\mbox{    }\mbox{    } 
  \texttt{cornelia@uic.edu} \\}

\begin{document}
\maketitle
\begin{abstract}
Existing Natural Language Inference (NLI) datasets, while being instrumental in the advancement of Natural Language Understanding (NLU) research, are not related to scientific text. In this paper, we introduce \textsc{SciNLI}, a large dataset for NLI that captures the formality in scientific text and contains $107,412$ sentence pairs extracted from scholarly papers on NLP and computational linguistics. Given that the text used in scientific literature differs vastly from the text used in everyday language both in terms of vocabulary and sentence structure, our dataset is well suited to serve as a benchmark for the evaluation of scientific NLU models. Our experiments show that \textsc{SciNLI} is harder to classify than the existing NLI datasets. Our best performing model with XLNet achieves a Macro F1 score of only $78.18\%$ and an accuracy of $78.23\%$ showing that there is substantial room for improvement. 


\end{abstract}

\section{Introduction}
Natural Language Inference (NLI) or Textual Entailment \cite{bowman-etal-2015-large} aims at recognizing the semantic relationship between a pair of sentences---whether the second sentence entails the first sentence, contradicts it, or they are semantically independent. NLI was introduced \cite{dagan2005pascal} to facilitate the evaluation of Natural Language Understanding (NLU) that significantly impacts the performance of many NLP tasks such as text summarization, question answering, 
and commonsense reasoning.
\vspace{1mm}

To date, several NLI datasets  are made available \cite{bowman-etal-2015-large, N18-1101, marelli-etal-2014-sick, dagan2005pascal}. These datasets have not only been instrumental for developing and evaluating NLI models but also have been useful in advancing many other NLP areas such as: representation learning \cite{conneau2017supervised},  transfer learning \cite{pruksachatkun-etal-2020-intermediate} and multi-task learning \cite{liu-etal-2019-multi-task}.
\vspace{1mm}

However, despite their usefulness, none of the existing NLI datasets is related to scientific text that is found in research articles. The vocabulary as well as the structure and formality used in sentences in scientific articles are very different from the sentences used in the everyday language. Moreover, 
the scientific text captured in research papers brings additional challenges and complexities not only in terms of the language and its structure but also the inferences that exist in it which are not available in the existing NLI datasets. 
For example, a sentence can present the reasoning behind the conclusion made in the previous sentence, while other sentences indicate a contrast or entailment with the preceding sentence. These inferences are crucial for understanding, analyzing, and reasoning over scientific work \cite{Luukkonen1992,kuhn2012structure,hall2008studying}. Therefore, ideally, the scientific language inference models should be evaluated on datasets which capture these inferences and the particularities seen only in scientific text. 

\setlength\dashlinedash{0.2pt}
\setlength\dashlinegap{1.5pt}
\setlength\arrayrulewidth{0.3pt}
\begin{table*}[t]
\centering
\small
  \begin{tabular} { p{6em} p{16em}  p{16em} p{6em}}
    \toprule
   \multicolumn{1}{c}{\rule{0pt}{2ex}\textbf{Class}} & \multicolumn{1}{c}{\rule{0pt}{2ex}\textbf{First Sentence}} & \multicolumn{1}{c}{\textbf{Second Sentence}}&
   \multicolumn{1}{c}{\rule{0pt}{2ex}\textbf{Linking Phrase}}\\[1ex] 
    \hline
    \textsc{Contrasting}  & Essentially, that work examines how a word gains new senses, and how some senses of a word may become deprecated. &  here we examine how different words compete to represent the same meaning, and how the degree of success of words in that competition changes over time. & `In contrast,'\\
    \hdashline
    \textsc{Reasoning}  & The Lang-8 corpus has often only one corrected sentence per learner sentence, which is not enough for evaluation. & we ensured that our evaluation corpus has multiple references. & `Thus,' \\
    \hdashline
    \textsc{Entailment}  & As a complementary area of investigation, a plausible direction would be to shift the focus from the decomposition of words into morphemes, to the organization of words as complete paradigms. & instead of relying on sub-word units, identify sets of words organized into morphological paradigms (Blevins, 2016).'' & `That is,'\\
    \hdashline
    \textsc{Neutral} & Literature on the topic of the current study spans across many areas, including verb classification, semiotics, sign language and learning. & abstract words can be more challenging to learn and memorise. & N/A \\
    \bottomrule
  \end{tabular}
  \caption{Examples of sentence pairs from our dataset and the linking phrases used to extract them, corresponding to all four classes considered. The second sentence of each pair is shown after removing the linking phrase.}
    \label{table:class_examples}
\end{table*}

\vspace{1mm}
To this end, we seek to enable deep learning for natural language inference over scientific text by introducing \textsc{SciNLI},\footnote{\url{https://github.com/msadat3/SciNLI}} a large dataset of $107,412$ sentence pairs extracted from scientific papers related to NLP and computational linguistics (CL) and present a comprehensive investigation into the inference types that occur frequently in scientific text. To capture the inference relations which are prevalent in scientific text but are unavailable in the existing NLI datasets, we introduce two new classes---\textsc{Contrasting} and \textsc{Reasoning}. 
We create \textsc{SciNLI} by harnessing cues in our data in the form of linking phrases between contiguous sentences, which are indicative of their semantic relations and provide a way to build a labeled dataset using distant supervision \cite{mintz-etal-2009-distant}. 
During training, we directly utilize these (potentially noisy) sentence pairs, but to ensure a realistic evaluation of the NLI models over scientific text, we manually annotate 6,000 sentence pairs. These clean pairs are used in two splits, 2,000 pairs for development and hyper-parameter tuning and 4,000 pairs for testing.
Table \ref{table:class_examples} shows examples from our dataset corresponding to all of our four classes.



\vspace{1mm}
We evaluate \textsc{SciNLI} by experimenting with 
traditional machine learning models using lexical and syntactic features, neural network models---BiLSTM, CBOW, CNN, and pre-trained language models---BERT \cite{devlin-etal-2019-bert}, SciBERT \cite{beltagy-etal-2019-scibert}, RoBERTa \cite{liu2019roberta}, and XLNet \cite{yang2019xlnet}. Our findings suggest that: (1) \textsc{SciNLI} is harder to classify than other datasets for NLI; (2) Lexical features are not enough for a model to achieve satisfactory performance on \textsc{SciNLI} and deep semantic understanding is necessary; (3) \textsc{SciNLI} is well suited for evaluating scientific NLI models; and (4) Our best performing model based on XLNet shows $78.18\%$ Macro F1 and $78.23\%$ accuracy illustrating that \textsc{SciNLI} is a challenging new benchmark.


\section{Related Work}
To date, several datasets exist for NLI of varying size, number of labels, and degree of difficulty. \citet{dagan06rte} introduced the RTE (Recognizing Textual Entailment) dataset of text-hypothesis pairs from the general news domain and considered two labels: entailment or no-entailment (i.e., a hypothesis is true or false given a text). The RTE dataset is paramount in developing and advancing the entailment task. The SICK (Sentences Involving Compositional Knowledge) dataset introduced by \citet{marelli-etal-2014-sick} was created from two existing datasets of image captions and video descriptions. SICK consists of sentence pairs (premise-hypothesis) labeled as: entailment, contradiction, or neutral. Despite being instrumental in the progress of NLI, both RTE and SICK datasets are less suitable for deep learning models due to their small size. 

In recent years, \textsc{SNLI} \cite{bowman-etal-2015-large} and \textsc{MNLI} \cite{N18-1101} are the most popular datasets for training and evaluating NLI models, in part due to their large size. Similar to \textsc{SICK}, \textsc{SNLI} is derived from an image caption dataset where the captions are used as premises and hypotheses are created by crowdworkers, with each sample being labeled as: entailment, contradiction, or neutral. \textsc{MNLI} is created in a similar fashion to SNLI except that the premises are extracted from sources such as face-to-face conversations, travel guides, and the 9/11 event, to make the task more challenging and suitable for domain adaptation. 
More recently, \citet{nie-etal-2020-adversarial} released \textsc{ANLI} which was created in an iterative adversarial manner where human annotators were used as adversaries to provide sentence pairs for which the state-of-the-art models make incorrect predictions.
Unlike the datasets specific to classifying the relationships between two sentences, \citet{zellers-etal-2018-swag} combined NLI with commonsense reasoning to introduce a new task of predicting the most likely next sentence from a number of options along with their new dataset called \textsc{SWAG} which was also created with an adversarial approach. However, different from \textsc{ANLI}, the \textsc{SWAG} approach was automatic. All these datasets have been widely used for evaluating NLU models and many of them appear in different NLU benchmarks such as \textsc{GLUE} \cite{wang-etal-2018-glue} and \textsc{SuperGLUE} \cite{wang2019superglue}.

\vspace{1mm}
Heretofore, \citet{khot2018scitail} created the only NLI dataset related to science. Their dataset, \textsc{SciTail} was derived from a school level science question-answer corpus. As a result, the text used in \textsc{SciTail} is 
very different from the type of text used in scientific papers. 
Furthermore, the sentence pairs in \textsc{SciTail} are classified into one of two classes: entailment or no-entailment. Thus, \textsc{SciTail} does not cover all the inference relationships necessary to understand scientific text.

\vspace{1mm}
In other lines of research, discourse cues, e.g., linking phrases have been previously used to extract inter-sentence and/or inter-clause semantic relations in discourse parsing \cite{hobbs1978discourse, webber1999discourse, prasad-etal-2008-penn, jernite2017discourse, nie-etal-2019-dissent}, causal inference \cite{do-etal-2011-minimally, radinsky2012learning, ijcai2020-0502, dunietz-etal-2017-corpus} and why-QA \cite{oh-etal-2013-question}. However, none of the aforementioned bodies of research investigates these relations in scientific text, nor do they exploit the discourse cues to create NLI datasets. Furthermore, discourse parsing studies a broader range of semantic relations, many of which are unrelated to the task of NLI while causal inference and why-QA are limited to only cause-effect relations.  In contrast to these tasks, we focus on the semantic relations which are either relevant to the task of NLI or highly frequent in scientific text and leverage linking phrases to create the first ever scientific NLI dataset, which we call \textsc{SciNLI}.

\section{\textsc{SciNLI}: A New Corpus for NLI}
In order to better understand the inter-sentence relationships that exist in scientific text, we started the process of creating our dataset by perusing through scientific literature with the intent of finding clues that are revealing of those relationships. We found that to have a coherent structure, authors often use different linking phrases in the beginning of sentences, which is indicative of the relationship with the preceding sentence. 
For example, to elaborate or make something specific, authors use linking phrases such as ``In other words'' or ``In particular,''  which indicate that the sentence supports or entails the previous sentence. We also found that some linking phrases are used to indicate additional relationships that are prevalent in scientific text but are not captured in the existing NLI datasets. For instance, when a sentence starts with ``Therefore'' or ``Thus,'' it indicates that the sentence is presenting a conclusion to the reasoning in the previous sentence. Similarly, the phrase ``In contrast'' is used to indicate that the sentence is contrasting what was said in the previous sentence. 

\vspace{1mm}
Therefore, inspired by the framework of discourse coherence theory \cite{hobbs1978discourse, webber1999discourse, prasad-etal-2008-penn} that characterizes the inferences between discourse units, we extend the NLI relations commonly used in prior NLI work—entailment, contradiction, and semantic independence—to a set of inference relations that manifest in scientific text—contrasting, reasoning, entailment, and semantic independence (\S\ref{sec:classes}). 
In order to create a large training set with minimal manual effort, we employ a distant supervision method based on linking phrases that are commonly used in scientific writing and are indicative of the semantic relationship between adjacent sentences (\S\ref{sec:data_collection}). We avoid the noise incurred by the distant supervision method in our development and test sets by manually annotating these sets (\S\ref{sec:eval_sets}). 


\subsection{Inference Classes} \label{sec:classes}
We define the inference classes used to create our dataset in this section.

\subsubsection{\textsc{Contrasting}}
Our \textsc{Contrasting} class is an extension of the \textsc{Contradiction} class in the existing NLI datasets. With this class, in addition to contradicting relations between sentences in a pair, we aim to capture inferences that occur when one sentence mentions a comparison, criticism, juxtaposition, or a limitation of something said in the other sentence. We can see an example of a sentence pair from our \textsc{Contrasting} class in Table \ref{table:class_examples}. Here, the authors discuss how their work differs from the other work mentioned in the first sentence thereby making a comparison between the two works. 

\subsubsection{\textsc{Reasoning}}
The examples where the first sentence presents the reason, cause, or condition for the result or conclusion made in the second sentence are placed in our \textsc{Reasoning} class. In Table \ref{table:class_examples}, we can see an example where the authors mention that they use a multi-reference corpus for evaluation in the second sentence and provide the reason behind it in the first sentence. 

\subsubsection{\textsc{Entailment}}
Our \textsc{Entailment} class includes the sentence pairs where one sentence generalizes, specifies or has an equivalent meaning with the other sentence. An example from this class can be seen in Table \ref{table:class_examples}. In the example, the second sentence is specifying the proposed direction mentioned in the first sentence
making the pair suitable for our \textsc{Entailment} class.

\subsubsection{\textsc{Neutral}}
The \textsc{Neutral} class includes the sentence pairs which are semantically independent. We can see an example from this class in Table \ref{table:class_examples}. Here, the first sentence discusses the span of the literature of a particular topic, whereas the second sentence mentions the challenges of handling abstract words in certain tasks. Therefore, the sentences are semantically independent of each other.

\begin{table}
\small
\begin{tabular}{lp{0.3\textwidth}}
\toprule
\textbf{Label} & \textbf{Linking Phrases}\\
\midrule
\textsc{Contrasting} & ‘However’, ‘On the other hand', ‘In contrast', ‘On the contrary'\\
 \midrule
\textsc{Reasoning} & ‘Therefore', ‘Thus', ‘Consequently’, ‘As a result', ‘As a consequence', ‘From here, we can infer’\\
\midrule
\textsc{Entailment} & ‘Specifically’, ‘Precisely’, ‘In particular’, ‘Particularly’, ‘That is’, ‘In other words’\\
\bottomrule
\end{tabular}
\caption{Linking phrases used to extract sentence pairs and their corresponding classes.
}
\label{table:linking_phrases}
\end{table}

\begin{table*}[t]
\centering
\small
  \begin{tabular}{ l r r r r r r r r r }
    \toprule
      &  \multicolumn{3}{c}{\bf \#Examples} & \multicolumn{2}{c}{\bf \#Words} & \multicolumn{2}{c}{\bf `S' parser} &  \\
       \cmidrule(lr){2-4}  \cmidrule(lr){5-6}  \cmidrule(lr){7-8}
   {\bf Dataset } & {\bf Train}     & {\bf Dev}   & {\bf Test} & {\bf Prem.} & {\bf Hyp.} & {\bf Prem.} & {\bf Hyp.} & {\bf Overlap} & {\bf Agrmt.}  \\ 
   \midrule
   \textsc{SNLI} & 550,152 & 10,000 & 10,000 & 14.1 & 8.3 & 74.0\% & 88.9\% & 52.97\% & 89.0\%\\  
   \textsc{MNLI} & 392,702 & 20,000 & 20,000 & 22.3 & - & 91.0\% & 98.0\% & - & 88.7\%\\
   \textsc{SICK} & 4,500 & - & 4,927 & 9.76 & 9.57 & - & - & 64.85\% & 84.0\%\\
   \textsc{SciTail} & 23,596 & 1,304 & 2,126 & 10.79 & 10.28 & 89.5\% & 99.1\% & 54.84\% & -\%\\ 
    \midrule 
    \textsc{SciNLI}  &  &  &  &  &  & & &  \\
    \mbox{{\color{white}{xxx}}+}\textsc{Contrasting} & 25,353 & 500 & 1,000 & 27.41 & 24.50 & 97.3\% & 97.4\% & 31.33\% & 91.6\%\\
    \mbox{{\color{white}{xxx}}+}\textsc{Reasoning} & 25,353 & 500 &  1,000 & 28.25 & 24.32 & 97.5\% & 97.7\% & 32.75\% & 74.6\%\\
    \mbox{{\color{white}{xxx}}+}\textsc{Entailment} & 25,353 & 500 & 1,000 & 27.08 & 28.90 & 96.9\% & 95.9\% & 32.98\% & 82.3\%\\
    \mbox{{\color{white}{xxx}}+}\textsc{Neutral} & 25,353 & 500 & 1,000 & 26.76 & 26.02 & 95.3\% & 95.6\% & 23.18\% & 94.7\%\\
    
\midrule
\textsc{SciNLI} Overall & 101,412 & 2,000 & 4,000 & 27.38 & 25.93 & 96.8\% & 96.7\% & 30.06\% & 85.8\%\\
    \bottomrule
  \end{tabular}
  \caption{Comparison of key statistics of \textsc{SciNLI} with other related datasets.}
    \label{table:data_stat_comparision_bal}
\end{table*}
\subsection{Training Set Creation}
\label{sec:data_collection}
We construct our training set from scientific papers on NLP and computational linguistics available in the ACL Anthology, published between 2000 and 2019 \cite{bird2008acl,radev-etal-2009-acl}. 
For extracting textual data from the PDF papers, we use GROBID\footnote{\url{github.com/kermitt2/grobid}} which is a popular tool for parsing PDF files. We employ the following distant supervision technique on the extracted text to select and label the sentence pairs.

\vspace{1.5mm}
We create a list of linking phrases which are indicative of the semantic relationship between the sentence they occur in and the respective previous sentence. 
We then group these linking phrases into three classes based on the type of relationship indicated by each of them. The linking phrases and their assigned class can be seen in Table \ref{table:linking_phrases}. We select the sentences which start with any of these phrases from each paper and include them in our dataset as hypotheses or second sentences; we include their respective preceding sentences as the premises or first sentences. 
Each sentence pair is labeled based on the class assigned to the linking phrase present in the second sentence, e.g., if the second sentence starts with ``In contrast'', the sentence pair is labeled as \textsc{Contrasting}. After assigning the labels, we delete the linking phrases from the second sentence of each pair to ensure that the models cannot get any clues of the ground truth labels just by looking at them. We also pair a large number of randomly selected sentences for our \textsc{Neutral} class using three approaches:
\vspace{-1mm}
\begin{itemize}
  \item \textsc{BothRand}: Two completely random sentences which do not contain any linking phrases are extracted (both from the same paper) and are paired together.
  \item \textsc{FirstRand}: First sentence is random; second sentence is selected randomly from the other three classes (both from the same paper).
  \item \textsc{SecondRand}: Second sentence is random; first sentence is selected randomly from the other three classes (both from the same paper).
\end{itemize} 
Our choice for including the last two approaches above was to make the dataset more challenging. 
\subsection{Benchmark Evaluation Sets Creation}
\label{sec:eval_sets}

To create our development and test sets, we start by extracting and labeling sentence pairs using the same distant supervision approach described in the previous section from the papers published in 2020 which are available in the ACL anthology. We then manually annotate a subset of these sentence pairs in order to make \textsc{SciNLI} a suitable benchmark for evaluation. The annotation process is completed in two steps, as described below.

\vspace{1.5mm}
First, we manually clean the data by filtering out the examples which contain too many mathematical terms and by completing the sentences that are broken due to erroneous PDF extraction by looking at the papers they are from. The second step of the annotation process is conducted in an iterative fashion. In each iteration, we randomly sample a balanced subset from the cleaned set of examples created in the previous step and present the sentence pair from each example to three expert annotators. To avoid a performance ceiling due to lack of context, the annotators are instructed to label each example based only on the two sentences in each example. If the label is not clear from the context available in the two sentences, the instruction is to label them as unclear. The label with the majority of the votes from annotators is then chosen as the gold label. No gold label is assigned to the examples ($\approx5\%$) which do not have a majority vote. The examples for which the gold label agrees with the label assigned based on the linking phrase are selected to be in our \textbf{benchmark evaluation set}. 
We continue the iterations of sampling a balanced set of examples and annotating them until we have at least $1,500$ examples from each class in the benchmark evaluation set. In total, $8,044$ sentence pairs---$2,011$ from each class are annotated among which $6,904$ have an agreement between the gold label and the label assigned based on the linking phrase. Therefore, these $6904$ examples are selected to be in the benchmark evaluation set. 
The percentage of overall agreement and the class-wise agreement between the gold labels and the labels assigned based on the linking phrases are reported in the last column of Table \ref{table:data_stat_comparision_bal}. The Fleiss-k score among the annotators is $0.62$ which indicates that the agreement among the annotators is substantial \cite{landis1977measurement}.

\vspace{-1mm}
We randomly select $36\%$ of the papers in our benchmark evaluation set to be in our development set and the rest of the papers are assigned to the test set. 
This is done based on our decision to have at least 500 samples from each class in the development set and 1000 samples from each class in the test set. Splitting the dataset into train, development and test sets {\em at paper level instead of sentence pair level} is done to prevent any information leakage among the data splits caused by sentences from one paper being in more than one split.

\subsection{Data Balancing}
Because of the differences in the frequency of occurrence of the linking phrases related to different classes, our initial dataset was unbalanced in all three splits. In contrast, the examples in the related datasets such as \textsc{SNLI} \cite{bowman-etal-2015-large} and \textsc{MNLI} \cite{N18-1101} are almost equally distributed across their classes. Therefore, 
for a fair comparison, we balance our dataset by downsampling the top three most frequent classes to the size of the least frequent class in each split. We can see the number of examples in each class of our \textsc{SciNLI} dataset in Table \ref{table:data_stat_comparision_bal}. 

\subsection{Data Statistics}
\vspace{-1mm}
A comparison of key statistics of \textsc{SciNLI} with four related datasets is also shown in Table \ref{table:data_stat_comparision_bal}. 

\paragraph{Dataset Size}
Although the total size of our dataset is smaller than \textsc{SNLI} and \textsc{MNLI}, \textsc{SciNLI} is still large enough to train and evaluate deep learning based NLI models.

\paragraph{Sentence Lengths}
From Table \ref{table:data_stat_comparision_bal}, we can see that the average number of words in both premise and hypothesis is higher in \textsc{SciNLI} compared with the other datasets. This reflects the fact that sentences used in scientific articles tend to be longer than the sentences used in everyday language. 

\vspace{-1mm}
\paragraph{Sentence Parses}
Similar to the related datasets, we parse the sentences in \textsc{SciNLI} by using the Stanford PCFG Parser (3.5.2) \cite{klein-manning-2003-accurate}. We can see that $\approx 97\%$ of both first and second sentences have parses with an `S' root which is higher than the sentences in \textsc{SNLI} and very competitive with the other datasets. This illustrates that 
most of our sentences 
are syntactically complete.

\vspace{-1mm}
\paragraph{Token Overlap}
We report the average percentage of tokens occurring in hypotheses which overlap with the tokens in their premises (Table \ref{table:data_stat_comparision_bal}). We observe that the overlap percentage in \textsc{SciNLI} is much lower compared to the other datasets. Therefore, our dataset has low surface-level lexical patterns revealing the relationship between sentences. 

\section{\textsc{SciNLI} Evaluation}
We evaluate our dataset by performing three sets of experiments. First, we aim to understand the difficulty level of \textsc{SciNLI} compared to related datasets (\S\ref{sec:lstm}). Second, we investigate a lexicalized classifier to test whether simple similarity based features can capture the particularities of our relations and potentially perform well on our dataset (\S\ref{sec:lex}). Third, we experiment with traditional machine learning models, neural network models and transformer based pre-trained language models to establish strong baselines (\S\ref{sec:baselines}). 

\setlength\dashlinedash{0.2pt}
\setlength\dashlinegap{1.5pt}
\setlength\arrayrulewidth{0.3pt}
\begin{table}[t]
\centering
\small
  \begin{tabular}{l c c }
    \toprule
{\bf Dataset} & {\bf F1}     & {\bf Acc} \\ 
   \midrule
   \textsc{SICK} & $63.54$ & $64.86$ \\
    \textsc{SNLI} & $80.61$ & $80.74$ \\
    \textsc{MNLI} Dev & & \\
    \mbox{{\color{white}{xxx}}-}Matched & $65.39$ & $65.70$ \\
    \mbox{{\color{white}{xxx}}-}Mismatched & $64.75$ & $65.01$ \\
    \textsc{SciTail} & $71.18$ & $72.29$ \\
    \textsc{SciNLI} & $60.98$ & $61.38$ \\
    \bottomrule
  \end{tabular}
  \vspace{-1.5mm}
  \caption{The Macro F1 (\%) and Accuracy (\%) of the BiLSTM model on different datasets. 
  }
  \vspace{-3mm}
    \label{table:lstm_results_clean}
    
\end{table}

\subsection{\textsc{SciNLI} vs. Related Datasets}
\label{sec:lstm}
To evaluate the difficulty of \textsc{SciNLI}, we compare the performance of a BiLSTM \cite{hochreiter1997long} based classifier on our dataset and four related datasets: \textsc{SICK}, \textsc{SNLI}, \textsc{MNLI} and \textsc{SciTail}. 
The architecture for this model is similar to the BiLSTM model used by \citet{N18-1101}. 
Precisely, the sentence level representations $S_1$ and $S_2$ are derived by sending the embedding vectors of the words in each of the sentences in a pair through two separate BiLSTM layers and averaging their hidden states. 
The context vector $S_c$ is calculated using the following equation:

\begin{equation}
\label{equation:combined_representation}
S_c = [S_1, S_2, S_1 \odot S_2, S_1 - S_2]
\end{equation}
Here, the square brackets denote a concatenation operation of vectors and $\odot$ and $-$ are element-wise multiplication and subtraction operators, respectively. $S_c$ is sent through a linear layer with Relu activation which is followed by a softmax layer to obtain the final output class.

\paragraph{Implementation details} We pre-process the input sentences by tokenizing and stemming them using the NLTK tokenizer\footnote{\url{https://www.nltk.org/api/nltk.tokenize.html}} and Porter stemmer,\footnote{\url{https://www.nltk.org/howto/stem.html}} respectively. Any stemmed token which occurs less than two times in the training set is replaced with an [UNK] token. We use 300D Glove embeddings \cite{pennington2014glove} to represent the tokens which are allowed to be updated during training. The hidden size for the BiLSTM models is 300. The batch size is set at 64 and the models are trained for 30 epochs where we optimize a cross-entropy loss using Adam optimizer \cite{kingma2014adam} with an initial learning rate of $0.001$. We employ early stopping with a patience size 10 where the Macro F1 score of the development set is used as the stopping criteria. Since \textsc{SICK} does not have a development split, we randomly select $10\%$ of its training examples to be used as the development set. Similarly, since \textsc{MNLI} does not have a publicly available test split, we consider its development split as the test split and we randomly select $\approx 10,000$ samples from the training set to be used as the development set.

We can see the performance of this model on different datasets in Table \ref{table:lstm_results_clean}. We find the following:

\paragraph{\textsc{SciNLI} is more challenging than other related datasets.}
The BiLSTM model shows a much lower performance for \textsc{SciNLI} compared with the other datasets. These results indicate that the task our dataset presents is more challenging compared to other datasets.  As we have seen in Table \ref{table:data_stat_comparision_bal}, there is a substantial amount of discrepancy in sentence lengths between \textsc{SciNLI} and the other datasets. The longer sentences in our dataset make it harder for the models to retain long distance dependencies, which result in lower performance. Furthermore, our dataset has low surface-level lexical cues and exhibits complex linguistic patterns that require a model to be less reliant on lexical cues but instead learn deep hidden semantics from text. 

\setlength\dashlinedash{0.2pt}
\setlength\dashlinegap{1.5pt}
\setlength\arrayrulewidth{0.3pt}
\begin{table}[t]
\centering
\small
  \begin{tabular}{ l c c c c }
    \toprule
&  \multicolumn{2}{c}{\bf SICK} & \multicolumn{2}{c}{\bf SciNLI}\\\cmidrule(lr){2-3} \cmidrule(lr){4-5}
   {\bf Features} & {\bf F1}     & {\bf Acc} & {\bf F1}     & {\bf Acc} \\ 
   \midrule
    {\textsc{Unigrams}} & 33.32  & 51.39 & 40.96 & 41.28 \\
    
    {\textsc{Bigrams}}  & 33.02 & 50.90  & 32.04 & 32.57\\
    {\textsc{Unigram \& Bigram}} & 34.52 & 49.69 & 39.35 & 39.52\\
    
    {\textsc{Features 1-3}}   & 66.68 & 71.86  & 35.75 & 38.15\\
     {\textsc{All Features}}  & 66.22 & 72.03  & 47.01 & 47.78\\
    \bottomrule
  \end{tabular}
  \vspace{-1.5mm}
  \caption{The Macro F1 (\%) and Accuracies (\%) of the lexicalized classifier on \textsc{SICK} and \textsc{SciNLI}. 
  }
    \label{table:lexicalized_results_clean}
    \vspace{-3mm}
\end{table}

\subsection{Lexical Similarity vs. Semantic Relationship}
\vspace{-1.5mm}
\label{sec:lex}
To verify that the examples in our dataset cannot be classified based only on syntactic and lexical similarities, we explore a simple lexicalized classifier similar to \cite{bowman-etal-2015-large}. We train a classifier using different combinations of the following features: (1) the second sentence's BLEU \cite{papineni2002bleu}
score with respect to the first sentence with an n-gram range of 1 to 4; (2) the difference in length between the two sentences in a pair; (3) overlap of all words, just nouns, verbs, adjectives, or adverbs - both the actual number and the percentage over possible overlaps; and (4) unigrams and bigrams from the second sentence as indicator features.
We compare the performance of these models on our dataset and the \textsc{SICK} dataset because given the small size of \textsc{SICK}, this is especially suitable for this kind of models. The results can be seen in Table \ref{table:lexicalized_results_clean}. We observe the following:

\paragraph{Semantic understanding is required to perform well on \textsc{SciNLI}.}
The lexicalized model fails to achieve satisfactory results on \textsc{SciNLI} even when all features are combined. Both Macro F1 and accuracy are much lower for our dataset than \textsc{SICK}. This means that without actually understanding the content in the sentences in \textsc{SciNLI}, a model cannot successfully predict their relationship. 

\subsection{\textsc{SciNLI} Baselines}
\label{sec:baselines}
To establish baselines on our dataset, we consider three types of models: a traditional machine learning model, neural network models, and pre-trained language models.

\paragraph{Traditional Machine Learning Model} We consider the lexicalized classifier using all four features described in \S\ref{sec:lex} as a baseline on our dataset.

\paragraph{Neural Network Models} We experiment with three neural models to get the sentence level representations for each sentence in a pair: (a) \textbf{BiLSTM} - word embeddings are sent through a BiLSTM layer and the hidden states are averaged; (b) \textbf{CBOW} - word embedding vectors are summed; (c) \textbf{CNN} - 64 convolution filters of widths [3, 5, 9] on the word embeddings are applied, the outputs of which are mean pooled to get a single vector representation from the filters of each of the three widths. These three vectors are then concatenated to get the sentence level representation. 

For all three models, the sentence level representations are combined as in Eq. \ref{equation:combined_representation}. 
The obtained representations are first sent through a linear layer with Relu activation followed by softmax for classification (i.e., project them with a weight matrix ${\bf W} \in \mathbb{R}^{d \times 4}$). The hyperparameters and other implementation details are the same as for the BiLSTM model described in \S\ref{sec:lstm}. 
\vspace{-1mm}

\paragraph{Pre-trained Language Models} We fine-tune four transformer based pre-trained language models: (a) \textbf{BERT} \cite{devlin-etal-2019-bert} - pre-trained by masked language modeling (MLM) on BookCorpus \cite{zhu2015aligning} and Wikipedia; 
(b)  \textbf{SciBERT} \cite{beltagy-etal-2019-scibert} - a variant of BERT pre-trained with a similar procedure but exclusively on scientific text; (c) \textbf{RoBERTa} \cite{liu2019roberta} - an extension of BERT which was pre-trained using dynamic masked language modeling, i.e.,  unlike BERT, different words were masked in each epoch during training. It was also trained for a longer period of time on a larger amount of text compared with BERT; and (d) \textbf{XLNet} \cite{yang2019xlnet} - pre-trained with a ``Permutation Language Modeling'' objective instead of MLM. We employ the base variants of each of these models using the huggingface\footnote{\url{https://huggingface.co/docs/transformers/}} transformers library. The input sequence for these models is derived by concatenating the two sentences in a pair with a \texttt{[SEP]} token in between. 
The 
\texttt{[CLS]} token is then projected with a weight matrix $\mathbf{W} \in \mathbb{R}^{d \times 4}$ by sending it as the input to a softmax layer to get the output class. We fine-tune each transformer based model for $5$ epochs where we minimize the cross-entropy loss using Adam optimizer \cite{kingma2014adam} with an initial learning rate of $2e-5$. Early stopping with a patience size 2 is employed.

The experiments are run on a single Tesla V10 GPU. The transformer based models took approximately four hours to train and the traditional machine learning and neural network models were trained in less than one hour. We run each experiment three times with different random seeds and report the average and standard deviation of the F1 scores for each of the four classes, their Macro average and overall accuracy in Table \ref{table:transformer_results_balanced_cleaned}.  Our findings are discussed below.

\setlength\dashlinedash{0.2pt}
\setlength\dashlinegap{1.5pt}
\setlength\arrayrulewidth{0.3pt}
\begin{table*}[t]
\centering
\small
  \begin{tabular}{ r c c c c c c }
    \toprule
&  {\textsc{Contrasting}} & {\textsc{Reasoning}} & {\textsc{Entailment}} & {\textsc{Neutral}} & {Macro F1} & Acc\\
   \midrule
    {Lexicalized} & $50.28 \pm 0.00$ & $37.18 \pm 0.00$ & $44.82 \pm 0.00$ & $55.77 \pm 0.00$ & $47.01 \pm 0.00$ & $47.78 \pm 0.00$ \\
    {CBOW} & $54.62 \pm 2.17$ & $50.54 \pm 1.75$ & $52.33 \pm 3.42$ & $49.25 \pm 0.18$ & $51.68 \pm 0.48$ & $51.78 \pm 0.53$ \\
    {CNN}  & $63.73 \pm 1.59$ & $58.86 \pm 1.17$ & $62.66 \pm 0.76$ & $56.40 \pm 0.97$ & $60.41 \pm 0.86$ & $60.53 \pm 0.85$ \\   
    {BiLSTM} & $63.93 \pm 0.53$ & $57.32 \pm 2.05$ & $64.01 \pm 0.56$ & $59.25 \pm 0.60$ & $61.12 \pm 0.15$ & $61.32 \pm 0.08$ \\
    \hline
    {BERT} & $77.46 \pm 0.30$ & $71.74 \pm 0.82$ & $75.09 \pm 0.13$ & $76.47 \pm 1.70$ & $75.19 \pm 0.35$ & $75.17 \pm 0.39$ \\
    \hline
    {SciBERT} & $80.30 \pm 0.60$ & $74.18 \pm 0.33$ & $75.90 \pm 1.47$ & $79.76 \pm 0.25$ & $77.53^{*} \pm 0.49$ & $77.52^{*} \pm 0.49$ \\
    {RoBERTa} & $81.18 \pm 0.77$ & $74.22 \pm 0.81$ & $77.99 \pm 0.52$ & $78.86 \pm 0.61$ & $78.06^{*} \pm 0.39$ & $78.12^{*} \pm 0.33$ \\
    {XLNet} & $81.53 \pm 0.30$ & $75.95 \pm 0.94$ & $77.63 \pm 0.38$ & $77.63 \pm 0.68$ & $\textbf{78.18}^{*} \pm \textbf{0.06}$ & $\textbf{78.23}^{*} \pm \textbf{0.12}$ \\ 
    \bottomrule
  \end{tabular}
  \caption{The Macro F1 scores (\%) and accuracies (\%) of our baseline models on \textsc{SciNLI} along with individual F1 scores on four classes. Here, an asterisk indicates that there is a statistically significant difference between the models in the third block of the table and BERT according to a paired T-test with $\alpha = 0.05$. The three models in the third block shows  statistically indistinguishable results. The best Macro F1 and accuracy are in bold.}
    \label{table:transformer_results_balanced_cleaned}
\end{table*}

\paragraph{Transformer based models consistently outperform the traditional models} The transformer based models have a very high performance gap with the traditional lexicalized and neural models. Their better performance can be attributed to their superior design for capturing the language semantics and their pre-training on large amounts of texts. 
\vspace{-3mm}

\vspace{-2mm}
\paragraph{More sophisticated pre-training methods lead to better performance}
RoBERTa and XLNet are created by addressing different limitations of BERT. Both of these models show a better performance than BERT on our dataset. Therefore, the progress made in these two models for better NLU capability is reflected by the results on \textsc{SciNLI}. This proves that \textsc{SciNLI} can be used as an additional resource for tracking the progress of NLU.


\paragraph{Pre-training on domain specific text helps to improve classification performance}
The results show that SciBERT consistently outperforms BERT on \textsc{SciNLI}. This is because unlike BERT, SciBERT was pre-trained exclusively on scientific text. Hence, 
it has a 
better capability to understand the text in the scientific domain. We see that RoBERTa and XLNet show slightly better performances than SciBERT despite being pre-trained on non-scientific text, just like BERT. However, it should be noted that these differences in performance are not statistically significant. Moreover, both RoBERTa and XLNet were created by modifying the training procedure of BERT to further improve the performance, whereas SciBERT is just a plain BERT model pre-trained on scientific text. Even without any modifications to the training procedure, SciBERT is able to perform similarly to these models proving the advantage of pre-training on domain specific text and suitability of our dataset for evaluating scientific NLI models.

\setlength\dashlinedash{0.2pt}
\setlength\dashlinegap{1.5pt}
\setlength\arrayrulewidth{0.3pt}
\begin{table}[t]
\centering
\small
  \begin{tabular}{ r r c c }
    \toprule
& &  \multicolumn{2}{c}{\bf \textsc{SciNLI}} \\  \cmidrule(lr){3-4} 
   {\bf Model} & & {\bf F1}     & {\bf Acc} \\ 
   \midrule
   BERT & \mbox{{\color{white}{xxx}}}\textsc{both sentences} & $75.36$ & $75.37$ \\
   & \mbox{{\color{white}{xxx}}}\textsc{only 2$^{nd}$ sentence} & $54.56$ & $55.40$ \\
   \midrule
   SciBERT &  \mbox{{\color{white}{xxx}}}\textsc{both sentences} & $77.66$ & $77.60$ \\
   & \mbox{{\color{white}{xxx}}}\textsc{only 2$^{nd}$ sentence} & $58.16$ & $58.80$ \\
    \bottomrule
  \end{tabular}
  \caption{Performance comparison on \textsc{SciNLI} when both sentences are concatenated vs. when only second sentence is used as the input.}
    \label{table:artifacts_balanced}
\end{table}

\section{Analysis}

Research has shown that some stylistic and annotation artifacts are present (only in the hypotheses) in NLI datasets created using crowdsource annotators  \cite{gururangan-etal-2018-annotation}. To verify that the models do not learn similar spurious patterns in our dataset and predict the labels without understanding the semantic relation between the sentences, we start our analysis by experimenting with only the second sentence as the input to BERT and SciBERT models. Next, to intuitively understand the errors made by the models, we perform a qualitative analysis of the predictions made by the SciBERT model on $100$ randomly selected examples from our test set. Finally, we show that the \textsc{Neutral} examples extracted with \textsc{FirstRand} and \textsc{SecondRand} approaches are harder to classify than the examples extracted with \textsc{BothRand}. 


\begin{table*}[t]
\centering
\small
  \begin{tabular} {p{16em}  p{16em} p{5.5em} p{5.5em}}
    \toprule
    \multicolumn{1}{c}{\rule{0pt}{2ex}\textbf{First Sentence}} & \multicolumn{1}{c}{\textbf{Second Sentence}}&
    \multicolumn{1}{c}{\rule{0pt}{2ex}\textbf{True Label}} &
   \multicolumn{1}{c}{\rule{0pt}{2ex}\textbf{Predicted Label}}\\[1ex] 
    \hline
    Multiple studies of BERT concluded that it is considerably overparametrized. & it is possible to ablate elements of its architecture without loss in performance or even with slight gains (Kovaleva et al., 2019;Michel et al., 2019;Voita et al., 2019). & \textsc{Entailment} & \textsc{Contrasting} \\
    \hline
     Upon further investigation, we find that experiments which use probabilities with image based features have an inter-quartile range of 0.05 and 0.1 for EBG and BLOG respectively whereas for experiments using probabilities with binning based features, this range is 0.32 for both datasets. &  inter-quartile range for experiments using ranks with image based features is 0.08 and 0.05 for EBG and BLOG whereas for experiments using ranks with binning based features, this range is 0.49 and 0.42 respectively. & \textsc{Contrasting} & \textsc{Neutral} \\
    \bottomrule
  \end{tabular}
  \caption{Examples of errors made by SciBERT on \textsc{SciNLI}. 
  }
    \label{table:error_analysis_clean}
\end{table*}


\paragraph{Spuriosity Analysis}
A comparison between the \textit{only second sentence} models and the models with both sentences concatenated as the input can be seen in Table \ref{table:artifacts_balanced}. Clearly, as we can see from the table, there is a substantial amount of performance decrease when only the second sentence is used as input. Therefore, in order to perform at the optimal level, both sentences are required for the models to make the correct inference by learning the semantic relation between them.




\paragraph{Qualitative Error Analysis}
We find that a major reason behind the wrong predictions is a lack of domain specific knowledge. For example, in the first sentence pair in Table \ref{table:error_analysis_clean}, without the domain knowledge that the number of parameters in a model affects the performance, one will not be able to make the correct inference. We also find that the model is prone to making mistakes for longer sentences. This issue is exemplified by the second sentence pair in Table \ref{table:error_analysis_clean}.

\paragraph{Neutral Class Performance Analysis}
We can see a plot of the accuracy shown by SciBERT on \textsc{Neutral} pairs of our test set extracted with different approaches in Figure \ref{figure:neutral_approach_vs_error_rate}. Indeed, the examples in which one sentence comes from one of the other three classes are harder to classify. 

 \definecolor{color1}{RGB}{96, 132, 240}
\definecolor{color2}{RGB}{46, 87, 209}
\definecolor{color3}{RGB}{15, 44, 128}
\begin{figure}[t]
\centering
\begin{tikzpicture}
\centering
\begin{axis}[
    width  = 5.5cm,
    height = 5cm,
    ticklabel style = {font=\small},
    xticklabel style={
    font=\small,
    rotate=18,
    },
    xlabel style={at={(axis description cs:0.5,-0.1)}},
    ymin=0, ymax=90,
    symbolic x coords={BothRand,SecondRand,FirstRand},
    ytick={0,20,40,60,80},
    xtick = {BothRand, SecondRand, FirstRand},
    xlabel={Extraction Approach},
    ylabel = {Accuracy (\%)},
    bar width=23pt,]
    \addplot[ybar,fill=color3] coordinates {
        (BothRand,83.95)
    };
    \addplot[ybar,fill=color2] coordinates {
        (SecondRand,77.90)
    };
    \addplot[ybar,fill=color1] coordinates {
        (FirstRand,68.97)
    };
\end{axis}
\end{tikzpicture}
\caption{Extraction approach vs. accuracy of SciBERT on the \textsc{Neutral} pairs of \textsc{SciNLI} test set.}
\label{figure:neutral_approach_vs_error_rate}
\end{figure}
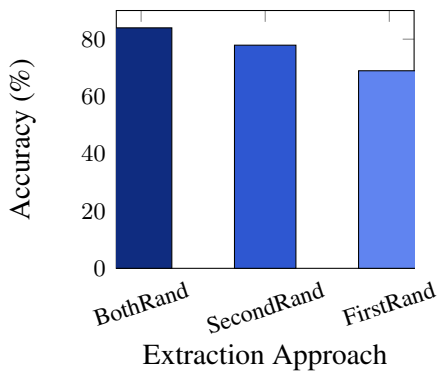

\section{Conclusion \& Future Directions}
In this paper, we introduced \textsc{SciNLI}, the first natural language inference dataset on scientific text created with our novel data annotation method. We manually annotated a large number of examples to create our benchmark test and development sets. Our experiments suggest that \textsc{SciNLI} is harder to classify than existing NLI datasets and deep semantic understanding is necessary for a model to perform well. We establish strong baselines and show that our dataset can be used as a challenging benchmark to evaluate the progress of NLU models. 
%
In the future, we will leverage knowledge bases to improve the models' ability to understand scientific text. 
We make our code and the \textsc{SciNLI} dataset available to further research in scientific NLI.

\section*{Acknowledgements}  This research is supported by NSF CAREER award 1802358 and NSF CRI award 1823292 to Cornelia Caragea. Any opinions, findings, and conclusions expressed here are those of the authors and do not necessarily reflect the views of NSF. We thank AWS for computing resources. We also thank our anonymous reviewers for their constructive feedback, which helped improve our paper.

\typeout{}
\bibliography{anthology, custom}
\bibliographystyle{acl_natbib}
\clearpage

\end{document}